\pdfoutput=1

\documentclass[11pt]{article}

\usepackage[final]{acl}

\usepackage{amsmath}
\usepackage{times}
\usepackage{latexsym}
\usepackage{mathabx}
\usepackage{array}

\usepackage[T1]{fontenc}

\usepackage[utf8]{inputenc}

\usepackage{microtype}
\usepackage{graphicx}
\usepackage{multirow}
\usepackage{multicol}
\usepackage{enumitem}
\usepackage{mathtools}
\usepackage{balance}
\newcommand{\mailtodomain}[1]{\href{mailto:#1@domain.com}{\nolinkurl{#1}}}

\title{Towards Personalized Intelligence at Scale}

\author{Yiping Kang$^1$\hspace{10pt} \textbf{Ashish Mahendra}$^2$\hspace{10pt} \textbf{Christopher Clarke}$^1$\hspace{10pt} \textbf{Lingjia Tang}$^1$\hspace{10pt} \textbf{Jason Mars}$^1$\\
\text{}$^1$\text{University of Michigan} \\
\text{}$^2$\text{BCS Technology}\\
\texttt{\{ypkang,csclarke,lingjia,profmars\}@umich.edu} \\
\texttt{\{ashish.mahendra\}@bcstechnology.com.au}}

\begin{document}
\maketitle
\begin{abstract}
\textit{Personalized Intelligence (PI)} is the problem of providing customized AI experiences tailored to each individual user.
In many applications, PI is preferred or even required~\cite{martinez2017personalized, rudovic2018personalized}.
Existing personalization approaches involve fine-tuning pre-trained models to create new customized models.
However, these approaches require a significant amount of computation to train, scaling with model size and the number of users, inhibiting PI to be realized widely.
In this work, we introduce a novel model architecture and training/inference framework to enable~\textit{Personalized Intelligence} at scale.
We achieve this by attaching a~\textit{Personalization Head (PH)} to pre-trained language models (LM). During training, the base LMs are frozen and only the parameters in PH are updated and are unique per user. This results in significantly smaller overall model sizes and training cost than traditional fine-tuning approaches when scaled across many users.
We evaluate PHs on academia and industry-focused datasets and show that the PHs outperform zeroshot baseline in F1 score and are significantly more scalable than traditional fine-tuning approaches.
We identify key factors required for effective PH design and training.

\end{abstract}

\section{Introduction}

As AI becomes ubiquitous in our lives, the experience remains largely homogeneous across users.
Companies invest a tremendous amount of data and time into training one or more sets of models, which are then served to all users~\cite{cost}.
However, in many cases, these "one-size-fits-all" models provide suboptimal experience on an individual level because of the high degree of heterogeneity in the user population \cite{anne-claire, measures_2021}.
This calls for \textit{Personalized Intelligence} enabled by AI models that can continuously learn and improve with user feedback and are tailored to each user.

\begin{figure}[ht]
    \centering
    \includegraphics[width=\columnwidth]{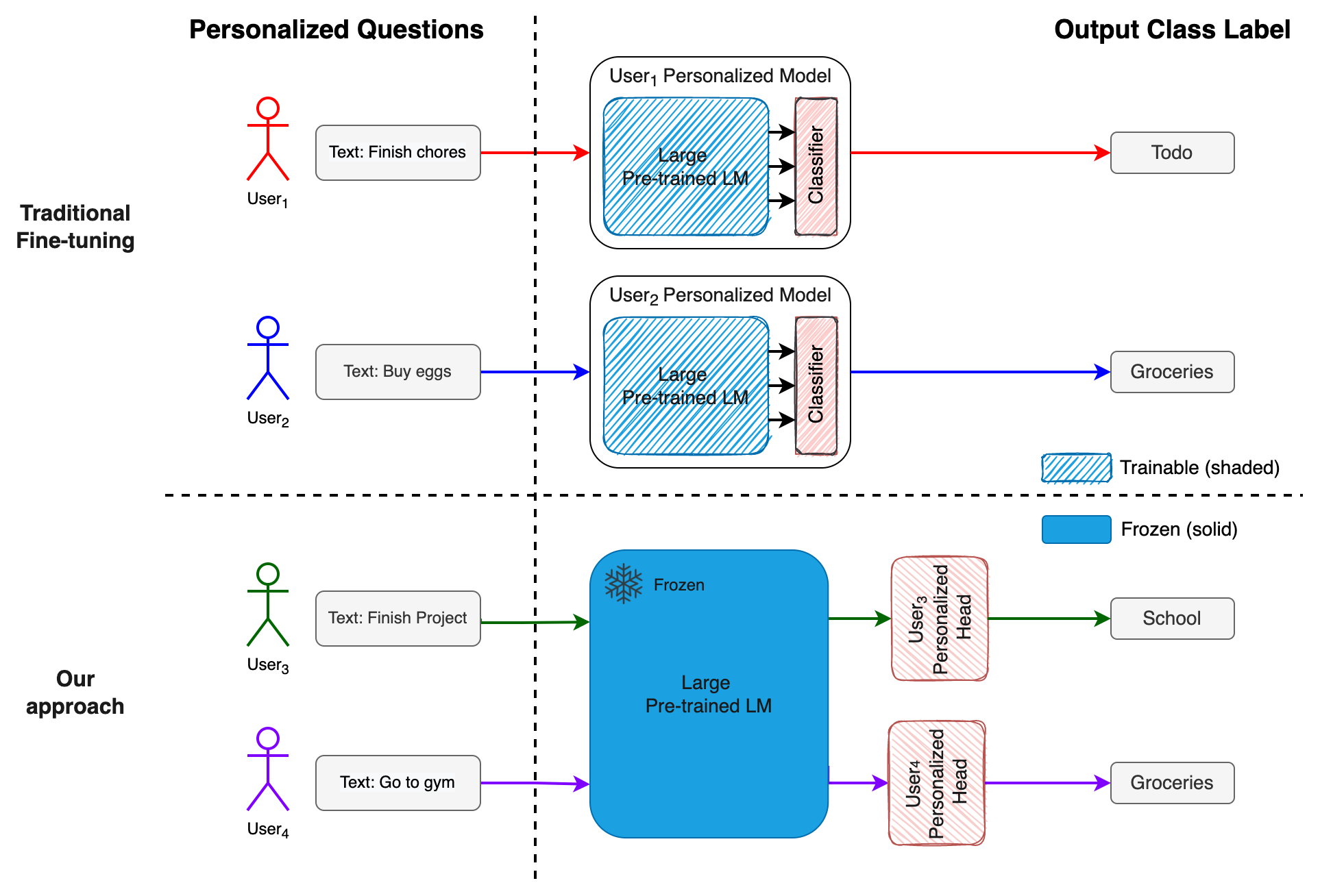}
    \caption{Example of individually personalized experience.}
    \label{fig:overview}
    \vspace{-1.8pc}
\end{figure} 

The major obstacle in realizing \textit{Personalized Intelligence} is the cost of training and storing individualized models in production \cite{sharir2020cost, strubell2019energy}.
State-of-the-art deep learning models are larger than ever and require significant computation cycles to train.
Therefore, the brute-force approach of fine-tuning pre-trained models for each individual user is not feasible due to the high compute and storage cost which scales with the number of users.

Zeroshot models have been studied recently as a method for applying pre-trained models to solve new problems without any additional training~\cite{gpt3, raffel2020exploring, sanh2021multitask}.
While the zeroshot approach is promising, there still exists a large performance gap in the level of accuracy that is needed for production usage~\cite{halder2020task, yinroth2019zeroshot}.
Thus, the research question remains as to how to enable~\textit{Personalized Intelligence} at a production scale for millions of users and beyond.

In this work, we propose a novel model training and inference framework for~\textit{Personalized Intelligence} at scale.
In order to enable individually personalized models for a large user population, we investigate the approach of attaching a small network called~\textit{Personalization Head (PH)} to a pre-trained language model.
In our framework, we train only the~\textit{PHs} while keeping the base models frozen, as shown in Figure~\ref{fig:overview}.
Our insight here is that the language representations from the state-of-the-art language models (e.g., BERT) capture high-level language features and can be "translated" to solve personalized problems.
We address the following key questions in this paper: 
\begin{enumerate}[topsep=1pt,itemsep=-1ex,partopsep=2ex,parsep=1ex]
    \item Can the PHs effectively leverage output from pre-trained LM despite not being trained jointly?
    \item How should PHs be designed to scale to millions of users and beyond?
    \item How should the PH design change based on its target personalization task?
\end{enumerate}

Transformer architectures have been shown to be effective in a wide range of translation tasks~\cite{vaswani2017attention, liu2020deep, gheini-etal-2021-cross}.
Leveraging this insight, we design our~\textit{PH} to be a single transformer encoder block with multi-head attention, followed by a linear classification layer.
Because the large-scale base language model is frozen during training, it can be used to serve across users and only the~\textit{PH} needs to be fine-tuned and stored per individual user.
We explore the design space of~\textit{PH} by experimenting with a spectrum of designs with varying model sizes and training costs.

For evaluation, we explore the efficacy of our~\textit{PHs} on academia and industry-oriented datasets, comparing with the performance of zeroshot models and traditional fine-tuning approach.
We show that the current zeroshot models are not effective in this context and the~\textit{PHs} outperform the zeroshot baselines by up to 61\% in F1 score.
Furthermore, we show that our personalization framework requires orders of magnitude less training cycles and model storage cost while achieving comparable performance when compared to fine-tuning the entire model.
We make the following specific contributions in this paper:
\begin{itemize}
    \item We introduce a novel training and inference framework for~\textit{Personalized Intelligence} where only a small~\textit{Personalized Head} needs to be trained and stored for each user instead of the traditional approach of fine-tuning the entire language model.
    \item We design~\textit{Personalization Heads (PH)}, a transformer-based module that effectively adapts output from pre-trained LMs to specific personalized problems without fine-tuning the base LM. The implementation of the proposed framework and PH architecture will be open-sourced.
    \item We evaluate the proposed PH on academia and industry-oriented datasets and show that it outperforms existing approaches in accuracy and scalability. We present a series of insights and design considerations for delivering personalized intelligence at scale. 
\end{itemize}

\section{Fine-tuning at Scale}
\newcommand{\argmax}[1]{\underset{#1}{\operatorname{arg}\operatorname{max}}\;}
\newcommand{\argmin}[1]{\underset{#1}{\operatorname{arg}\operatorname{min}}\;}
Let $M$ be a pre-trained language model (LM) and $\Theta_M$ be the set of parameters of $M$.
When applying $M$ to a downstream task $T$ with labelled training data $D_T$, a linear output layer $L$ with parameters $\Theta_L$ is attached to $M$.
$M$ and $L$ are trained jointly:
\begin{equation}\label{eq:traditiona-ft}
    \Theta\prime_M, \Theta\prime_L\gets \argmin{\Theta_M, \Theta_L} L_T(D_T;\Theta_M, \Theta_L)
\end{equation}
where $L$ is the loss function and $\Theta\prime_M$ and $\Theta\prime_L$ are the fine-tuned parameters of the language model $M$ and linear layer $L$, which are distinct from $\Theta_M$.
Let $\Omega(\Theta)$ be the computation complexity required to train a set of parameters $\Theta$.
We then define the training complexity of the above fine-tuning operation as:
\begin{equation}
    \Omega(\Theta\prime_M) + \Omega(\Theta\prime_L)
\end{equation}

Consider the scenario of fine-tuning for an individual user to create a personalized model.
We define $U=\{(D_1,L_1),(D_2, L_2),...,(D_N,L_N)\}$ as the collection of personalization tasks where $i\in[1,...,N]$ for $N$ users.
$D_i$ is the unique data for user $i$ and $L_i$ is the loss function.
The problem of fine-tuning to personalize for user $i$ becomes
\begin{equation}
    \Theta\prime_{M,i}, \Theta\prime_{L,i}\gets \argmin{\Theta_M, \Theta_L} L_i(D_i;\Theta_M, \Theta_L)
\end{equation}
The aggregated training complexity is
\begin{equation}\label{eq:scaling}
    \sum_1^N\Omega(\Theta\prime_{M,i}) + \sum_1^N\Omega(\Theta\prime_{L,i})
\end{equation}
The collection of model parameters to be stored is:
\begin{equation}
    \sum_1^N\vert\Theta\prime_{M,i}\vert + \sum_1^N\vert\Theta\prime_{L,i}\vert
\end{equation}

\label{subsec:head-details}
\begin{figure*}[ht]
    \centering
    \includegraphics[width=2\columnwidth]{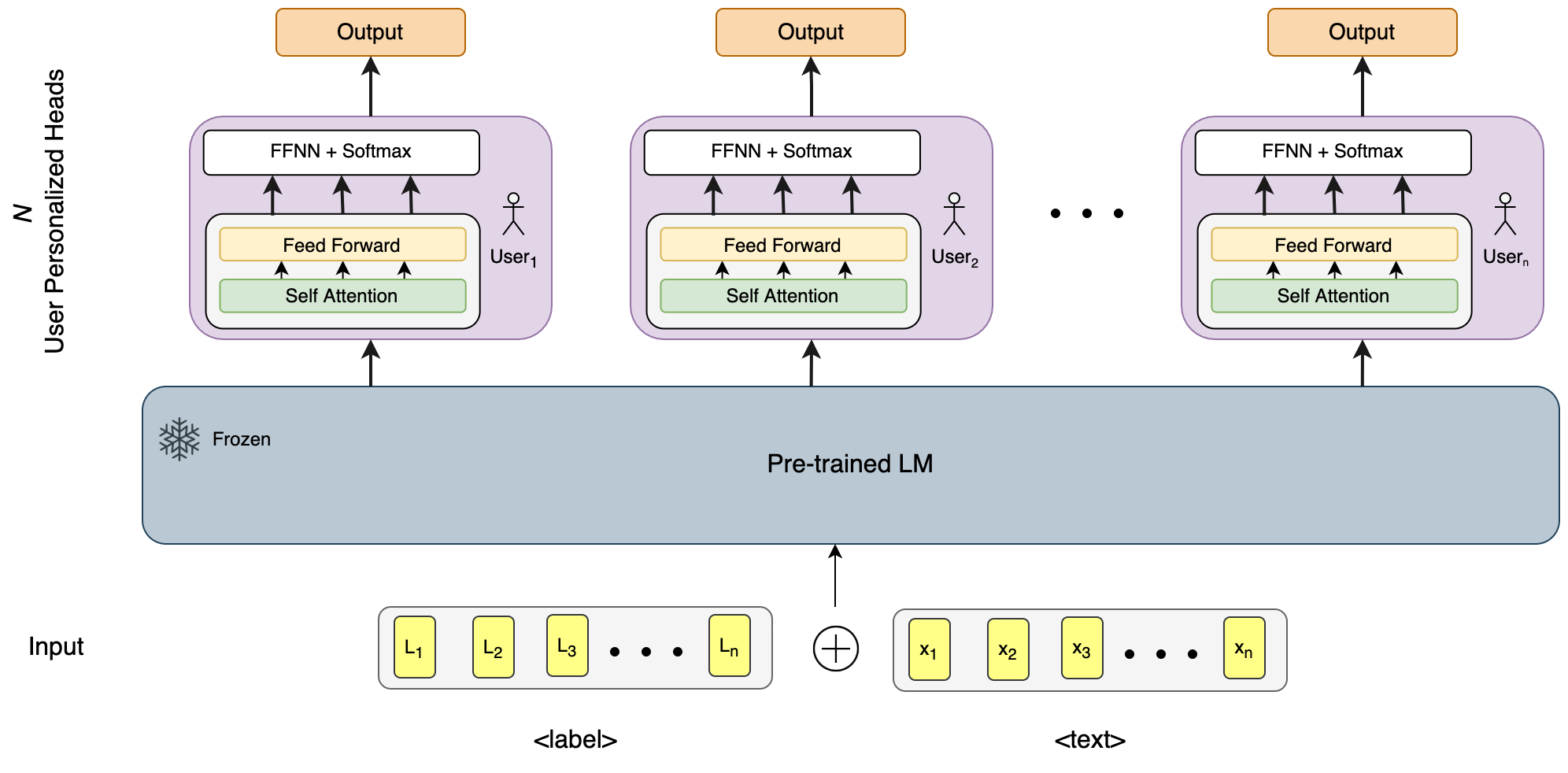}
    \caption{Overview of the proposed approach for personalization at scale with Personalization Head (PH).}
    \label{fig:arch}
\end{figure*} 

\section{Scalable Personalized Intelligence}
Large-scale pre-trained language models can be adapted and personalized for each individual through the fine-tuning process defined above.
The shortcoming is LMs ($M$) can have millions to billions of parameters ($\Theta_M$)~\cite{brown2020language}.
As a result, the training complexity ($\Omega(\Theta_M)$) is extraordinarily high as it scales with the number of users, as defined in Equation~\ref{eq:scaling}.

Inspired by the effectiveness of adapting pre-trained LMs to new tasks, we hypothesis that by adding a lightweight transformer module between the LM and the linear output layer, we can eliminate the need for fine-tuning the LMs and keep it constant across all users.
The intuition is that the LMs learn to capture generalized language features and a small transformer can leverage that for new tasks.
To this end, we introduce~\textit{Personalization Head (PH)}.
Figure~\ref{fig:arch} provides an overview of the proposed approach.

\subsection{Fine-tuning with PH}
We define $PH_i$ as the Personalization Head for user $i$.
When fine-tuning for a given user, we keep the LM parameters frozen and only train the PH and the linear output layer:
\begin{equation}
     \Theta\prime_{PH}, \Theta\prime_L\gets \argmin{\Theta_{PH}, \Theta_L} L_T(D_T;\Theta_M, \Theta_PH,  \Theta_L)
\end{equation}
Note that, compared to the traditional fine-tuning defined in Equation~\ref{eq:traditiona-ft}, no $\Theta\prime_{M}$ is generated.
The aggregated training complexity for fine-tuning with PH is:
\begin{equation}
    \sum_1^N\Omega(\Theta\prime_{PH,i}) + \sum_1^N\Omega(\Theta\prime_{L,i})
\end{equation}
and the total parameters is:
\begin{equation}
    \Theta_{M} + \sum_1^N\vert\Theta\prime_{PH,i}\vert + \sum_1^N\vert\Theta\prime_{L,i}\vert
\end{equation}
PH's model architecture is much smaller than a state-of-the-art language model. Therefore the model size and training cost is significantly lower:
\begin{equation}
    \vert\Theta_{PH}\vert<<\vert\Theta_M\vert{,\ \ }\Omega(\Theta\prime_{PH})<<\Omega(\Theta\prime_{M})
\end{equation}
As a result, the training complexity and model size are significantly reduced when fine-tuning with PH.

\subsection{Universal Binary Classification}
We aim to design a personalization framework that is generalizable to arbitrary classification tasks without requiring modification to the model architecture.
To that end, we draw inspiration from~\cite{halder2020task} and formulate the multi-class classification problem as a series of binary classification tasks:
\begin{equation}
   f(label(y_{i}),x) = P(True|y_{i},x) \, \forall \,Y
\end{equation}
We provide the model with both the class label $label(y_i)$ and the input text $x$ and the output layer generates a binary $True/False$ prediction with a confidence score $P$.
The class with the most confident $True$ prediction is selected as the classification prediction:
\begin{equation}
    y=\argmax{i\in\{1...M\}}f(label(y_i), x)
\end{equation}
where $M$ is the number of classes.

\subsection{PH Architecture}
\label{subsec:ph-architecture}
The functionality of the PH is converting the feature representation from pre-trained language models to new representations suited for a new problem unique to an individual. 
This is akin to a language translation problem, at which transformer models have been proven to perform well~\cite{raganato2018analysis,wang2019learning,lu-etal-2021-attention}.
We draw inspiration from these works and base our PH design on the Transformer architecture.

Figure~\ref{fig:arch} shows an overview of the PH, which consists of a single Transformer encoder layer.
We follow the Transformer architecture defined in the original transformer paper~\cite{vaswani2017attention}.
Each PH has a multi-head self-attention layer and two fully connected layers, followed by layer normalization~\cite{ba2016layer}.
Dropout~\cite{srivastava2014dropout} is applied to the output of the fully connected layers.

It is important to understand how to configure the encoder block to design an effective PH.
To help us explore the design space, we parameterize the following configurations: the size of the hidden dimension of the feed-forward network in the encoder and the number of attention heads in the attention layer.
We select these parameters to study because they have a significant effect on the size and capability of a Transformer encoder.
We investigate the impact of these design decisions in detail in Section~\ref{sec:results}.

\section{Experiments}
\begin{table}
\resizebox{\columnwidth}{!}{%
\begin{tabular}{c|c|c|c|c}
    \textbf{Dataset} & \textbf{Description} & \textbf{\# Classes} & \textbf{\# Train} & \textbf{\# Test} \\ \hline
    SNIPS & Smart assistants questions & 7 & 13,034 & 1,442 \\
    Clinc150 & Production VA tasks & 150 & 15,100 & 1,500 \\ \hline
\end{tabular}}
\caption{Dataset Statistics}
\label{tab:dataset}
\end{table}

\subsection{Datasets}
\label{subsec:datasets}
We evaluate the proposed approach by applying PHs to a pre-trained LM to the SNIPS dataset~\cite{coucke2018snips} and Clinc-150 dataset~\cite{larson-etal-2019-evaluation}.

We select SNIPS because its intents cover many common classification topics and it is a representative dataset widely studied in the literature.
We select Clinc-150 for its focus on the complexity of production use cases.
It has 150 intents and features intents and sentences inspired by real virtual assistants in production.
An overview of the datasets is shown in Table~\ref{tab:dataset}.
Due to the large number of testing examples in Clinc's original test set, we randomly sample 10 examples (out of 30) per class to construct a representative test set.
This test set is used across all experiments, including baselines and PHs.

We generate $<class, text>$ pair as training and testing input and $True/False$ label as output for each example in the dataset.
We include only the $True$ examples and leave out the $False$ examples to optimize training time, as training with $False$ examples increases training time by a factor equal to the number of possible classes~\cite{halder2020task} (7 for SNIPS and 150 for Clinc-150)

This issue is further exacerbated in production use cases, where there can exist hundreds of classes, as in the Clinc-150 dataset.

\begin{table*}[t]
\small
\centering\scriptsize
\setlength\extrarowheight{2pt}
\begin{tabular}{c|c|c|c|c||c|c} 
\multicolumn{3}{c|}{\textbf{Model}} & \textbf{\shortstack{\# Params\\/ User}} & \textbf{\shortstack{Size\\/ User}} & \shortstack{\textbf{SNIPS}\\F1 / Acc.} & \shortstack{\textbf{Clinc}\\F1 / Acc.} \\
                                    \hline 
\multirow{2}{*}{ZeroShot} & \multicolumn{2}{|c|}{BERT} & NA & NA & 2.99 / 1.60 & 0.78 / 0.47 \\
                                    & \multicolumn{2}{|c|}{TARS} & NA & NA & 35.27 / 26.70 & 23.98 / 23.67 \\
                                    \hline
\multirow{2}{*}{\shortstack{Fine-tuning \\LM + Linear Layer}} & \multicolumn{2}{c|}{BERT} & 109M & 417MB & 98.61 / 98.61 & 95.74 / 95.07\\
                             & \multicolumn{2}{c|}{TARS} & 109M & 417MB & 98.13 / 98.06 & 95.22 / 94.27 \\
                             \hline
\multirow{2}{*}{\shortstack{Fine-tuning \\Linear Layer Only}} & \multicolumn{2}{c|}{BERT} & 1.5K & 7KB & 68.70 / 58.67 & 52.43 / 50.27\\
                             & \multicolumn{2}{c|}{TARS} & 1.5K & 7KB & 71.12 / 63.11 & 33.27 / 33.20 \\
                             \hline
\multirow{15}{*}{\shortstack{Personalization Head (PH)\\ w/ frozen LM}} & \textbf{Hidden Dim} & \textbf{\# Attn. Heads} & & & & \\
\cline{2-7}
                                & \multirow{3}{*}{2048} & 8 & \multirow{3}{*}{5.52M} & \multirow{3}{*}{21MB} & 96.52 / 96.12 & 76.36 / 75.93 \\
                                &                       & 4 & & & 97.18 / 96.74 & 76.46 / 76.47 \\
                                &                       & 2 & & & 97.33 / 97.16 & 76.36 / 75.93 \\
                                \cline{2-7}
                                & \multirow{3}{*}{1024} & 8 & \multirow{3}{*}{3.94M} & \multirow{3}{*}{15MB} & 97.46 / 97.09 & 75.07 / 74.40 \\
                                &                       & 4 & & & 96.76 / 96.39 & 75.82 / 75.73 \\
                                &                       & 2 & & & 96.77 / 96.67 & 76.61 / 76.60\\
                                \cline{2-7}
                                & \multirow{3}{*}{512} & 8 & \multirow{3}{*}{3.15M} & \multirow{3}{*}{12MB} & 97.05 / 97.02 & 75.95 / 76.33 \\
                                &                       & 4 & & & 96.26 / 95.49 & 70.79 / 71.20 \\
                                &                       & 2 & & & 96.94 / 96.74 & 75.29 / 75.27 \\
                                \cline{2-7}
                                & \multirow{3}{*}{256} & 8 & \multirow{3}{*}{2.76M} & \multirow{3}{*}{11MB} & 95.90 / 95.70 & 66.99 / 67.53 \\
                                &                       & 4 & & & 95.64 / 95.15 & 68.64 / 67.87 \\
                                &                       & 2 & & & 97.06 / 96.32 & 67.68 / 67.13 \\
                                \cline{2-7}
                                & \multirow{3}{*}{128} & 8 & \multirow{3}{*}{2.57M} & \multirow{3}{*}{9.8MB} & 95.32 / 95.28 & 63.43 / 64.53 \\
                                &                       & 4 & & & 96.32 / 96.32 & 62.70 / 63.67 \\
                                &                       & 2 & & & 96.36 / 96.36 & 63.82 / 64.60 \\ \hline

\end{tabular}
\caption{F1 score and accuracy of the proposed PHs and baselines of zeroshot, fine-tuning the LM + linear head and fine-tuning the linear head only, evaluated on SNIPS and Clinc-150 datasets. We also show the number of parameters that are required to be fine-tuned for each approach, as well as the size of the personalized model to be managed for each user as a representation of the scalability of each approach.}
\label{tab:all-f1}
\end{table*}

\begin{table*}[h]
\centering\scriptsize
\setlength\extrarowheight{2pt}
\begin{tabular}{c|c|c|c|c|c|c}
\textbf{Hidden Dims} & \textbf{\# Params} & \textbf{Size} & \shortstack{\textbf{epoch=50}\\\textbf{\# data/class=100}} & \shortstack{\textbf{epoch=100}\\\textbf{\# data/class=100}} & \shortstack{\textbf{epoch=50}\\\textbf{\# data/class=200}} & \shortstack{\textbf{epoch=100}\\\textbf{\# data/class=200}} \\ \hline
2048 & 2.7M & 10.5MB & 65.92 & 70.43 (+4.50) & 78.12 (+12.20) & 92.99 (+27.70) \\
1024 & 3.2M & 12.0MB & 57.42 & 65.39 (+7.97) & 80.17 (+22.75) & 93.72 (+36.30) \\
512 & 3.9M & 15.0MB & 61.75 & 67.01 (+5.26) & 83.72 (+21.97) & 93.99 (+32.24) \\
256 & 5.5M & 21.0MB & 57.40 & 63.58 (+6.18) & 79.85 (+22.45) & 94.58 (+37.18) \\ \hline
\end{tabular}
\caption{F1 score (and differential) with increasing training data and/or training for more epochs}
\label{tab:data-vs-epoch}
\end{table*}

\subsection{Experimental Settings}
We implement our model using the Flair NLP framework~\cite{akbik2019flair} with an underlying pytorch runtime~\cite{NEURIPS2019_9015}.
We use the uncased BERT encoder as the base LM~\cite{devlin2018bert}.
We train for 50 epochs (unless noted otherwise) and report F1 scores on the test set.
For hyper-parameters, we use a batch size of 16 and a learning rate of 0.02, following the standard in ~\cite{halder2020task}.

\section{Results}
\label{sec:results}
\label{subsec:results}
We compare PHs with fine-tuning both LM and linear output layer, fine-tuning the linear layer only, and applying the model zeroshot.

\subsection{Understanding zero-shot efficacy}
We first investigate the efficacy of the zeroshot approach.
We use the TARS zeroshot classifier from~\cite{halder2020task} to predict the test examples without additional training and measure the F1-score.
The TARS classifier uses BERT as the underlying language model and is pre-trained on a suite of datasets including classification datasets such as AGNews and DBPedia~\cite{Zhang2015CharacterlevelCN}.
We test the zeroshot performance of both the trained TARS classifier as well as BERT out-of-the-box, as shown in the first row of table~\ref{tab:all-f1}.
BERT (out-of-the-box) achieves F1-score of 2.99 and 0.78 on SNIPS and Clinc, respectively, and the TARS classifier achieves an F1-score of 35.27 and 23.98 on SNIPS and Clinc, respectively, which are significantly lower than the reported state-of-the-art results.
This shows that zeroshot approach requires significant improvement to reach the level of performance needed for production usage.

\subsection{PH performance}
\label{subsec:all-together}

Table~\ref{tab:all-f1} shows the performance of the Personalization Head (PH) on SNIPS and Clinc.
We experiment with a wide range of PH configurations by varying the hidden dimension of the feed-forward layer and the number of attention heads in the PH.
For brevity, we include results for configurations with hidden dimensions from 128 to 2048 and \# attention heads of 2, 4, and 8.
Results for additional configurations are included in the Appendix.
We compare to the current approach of fine-tuning both LM and the linear output layer, and fine-tuning only the linear output layer while keeping the base LM frozen.
We also include in Table~\ref{tab:all-f1} the number of training parameters and model size required per user for PHs and the baselines.

We observe that PHs, across all configurations, significantly outperform fine-tuning only the linear layer on both datasets.
When comparing to the baseline of fine-tuning the entire model stack of the base language model and the linear output layer, PH achieves similar results for the SNIPS dataset, while requiring orders of magnitude less training cost.
Fine-tuning both LM and linear layer achieves higher F1-score on Clinc than PHs but it requires, for each user, the training of 109 million parameters which generates a 417MB model, while each PH only incurs for each user 2-5 million additional parameters to learn and 11-21MB model to store.
The large computation cycles and storage capacity required for the fine-tuning approach renders it inapplicable for production use cases that require scaling to many users.
We analyze the scalability impact of PHs in more detail in Section~\ref{subsec:scalability}

\section{Analysis}

\begin{figure*}[h]
    \small
    \centering
        \includegraphics[width=1.9\columnwidth]{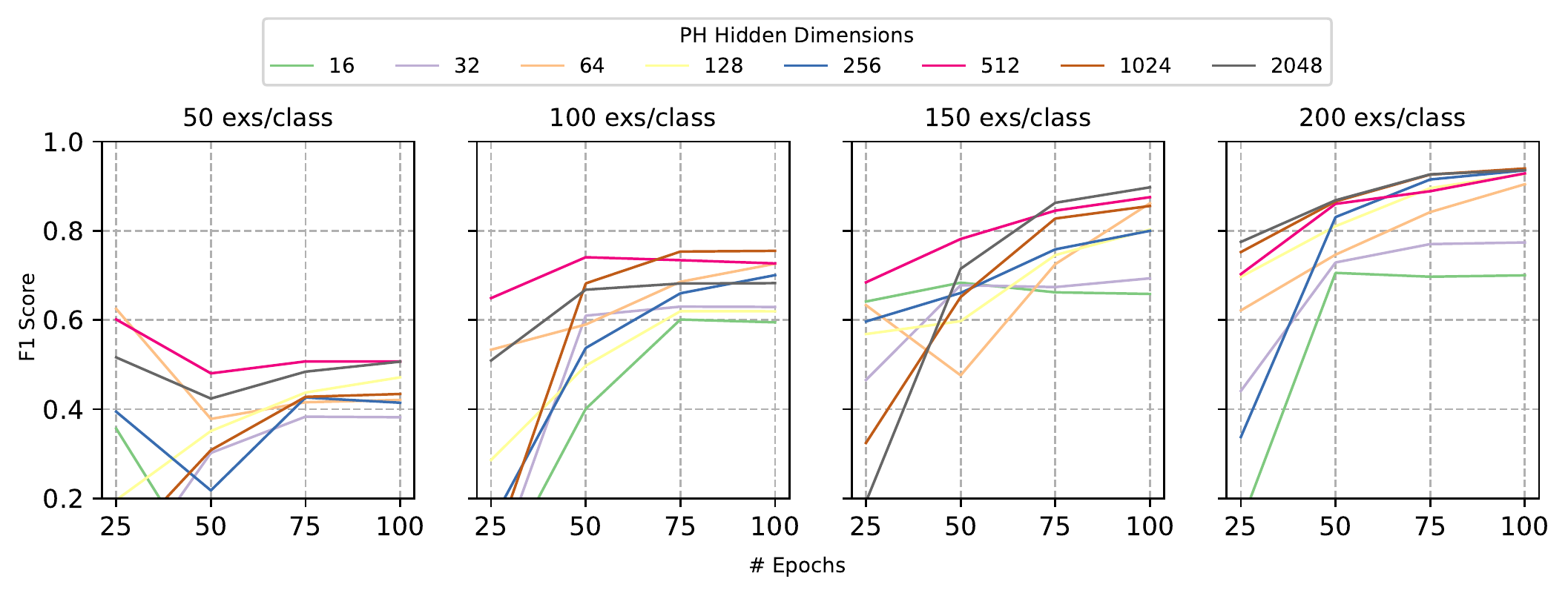}
        \caption{F1 Score w.r.t \# training epochs, for fixed amounts of data.}
        \label{fig:constant-datapoint}
\end{figure*} 

\begin{figure*}[h]
    \small
    \centering
        \includegraphics[width=1.9\columnwidth]{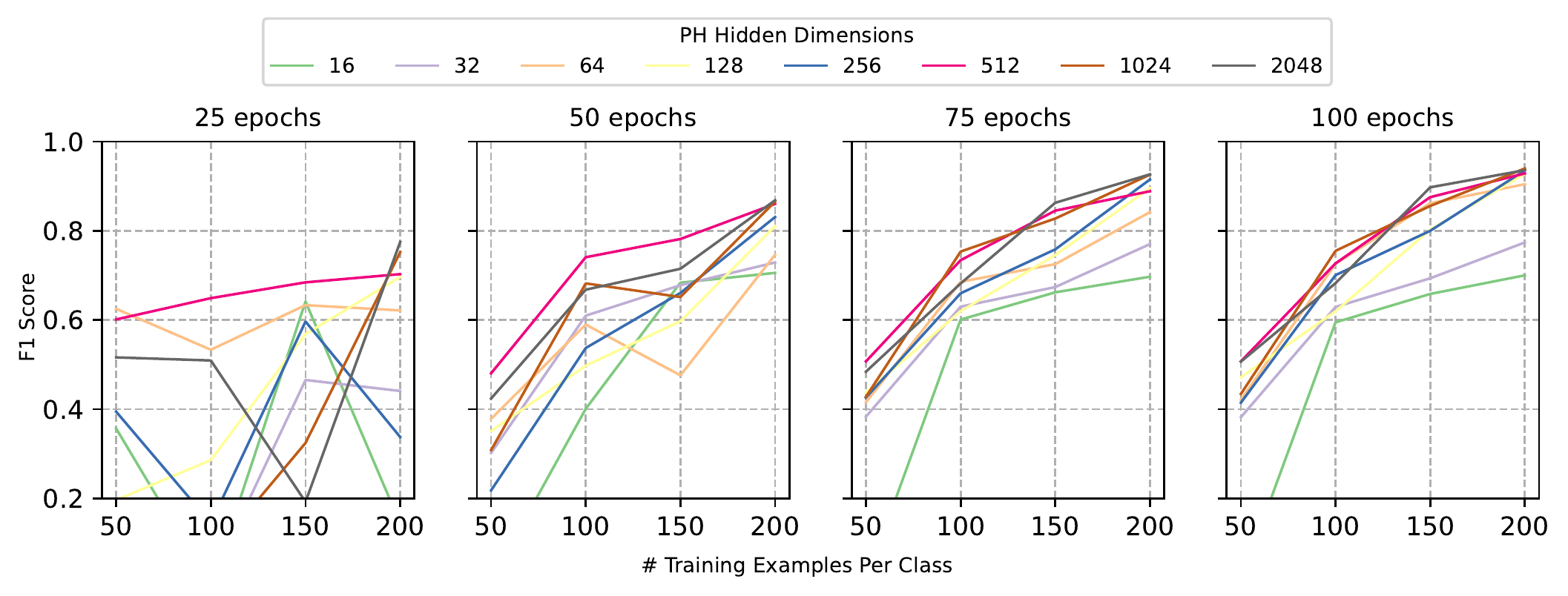}
        \caption{F1 Score w.r.t \# training data, for fixed amounts of epochs.}
        \label{fig:constant-epoch}
\end{figure*} 
We conduct experiments aimed at understanding the learning behavior of PHs and gain insights into how to design and deploy an effective PH for real-world use cases.
We aim to answer the following questions:
1) How to effectively train PHs in production?
2) How does the PH configuration affect its learning behavior?
3) Do larger PHs achieve better performance and does there exist a sweet spot of PH design that is the most compute and data efficient?

\subsection{Impact of Data vs. Epoch on Training PHs}
\label{subsec:data-vs-epoch}
We study the impact of the scale of training data and the number of training epochs on the PH performance.
In real-time training in production, there is often limited training data and training cycles available.
Therefore, it is imperative to understand how to train a PH in a compute and data efficient manner.
To this end, we construct a SNIPS sub-dataset by random sampling 100 training examples per class (1400 samples in total) and keep the full SNIPS test set.
We train the spectrum of PH designs for 50 epochs and then 50 more epochs (100 epochs in total) and record the test set F1-score at both points.
We then select another 100 training samples per class to add to the training set and repeat the same experiment.
Table~\ref{tab:data-vs-epoch} shows the average F1 scores, as well as improvement gained by increasing training data, training epochs, and both.

We observe that increasing the training data from 100 per class to 200 per class provides a significantly higher F1 score increase (+19.85 on average), compared to training for more epochs (+5.98 average).
This behavior is consistent across the various PH configurations.
This is intuitive because 100 training samples/class represents only 5.3\% of the full SNIPS training set and does not provide robust coverage of the problem space.
Increasing the training data scale should be the priority over more training iterations in the early stages of applying a PH to a personalized problem.

\subsection{PH Design Analysis}

Two main design choices for PHs are the hidden dimension of the encoder block and the number of attention heads in the multi-attention layer.
We study how these design choices impact the learning behavior of the PH.
We conduct a set of experiments where we gradually increase the amount of training data or epochs and measure the F1-score at each stopping point. 
This is to simulate a training setup in production, where the model gradually gets exposed to more training data as the applications collect more personalized data from the users.

\textbf{Hidden Dimension Size} Figure~\ref{fig:constant-datapoint} shows the F1 score of PHs with different hidden dimensions as they are trained with more epochs on the same amount of training data. 
We experiment with 50, 100, 150, and 200 training examples per class for 25, 50, 75, and 100 epochs.
Conversely, Figure~\ref{fig:constant-epoch} shows the F1 score of the same suite of PHs as they are trained with more training examples for the same number of training epochs.
We make several observations.

First, we observe that when exposure to training data is limited in the early stages of training, PH training can exhibit unpredictable behavior.
This is shown in the leftmost graphs of Figure~\ref{fig:constant-datapoint} and~\ref{fig:constant-epoch}, where the model performance is not improved with additional training data or more training epochs.

\begin{figure}[h]
    \centering
        \includegraphics[width=0.85\columnwidth]{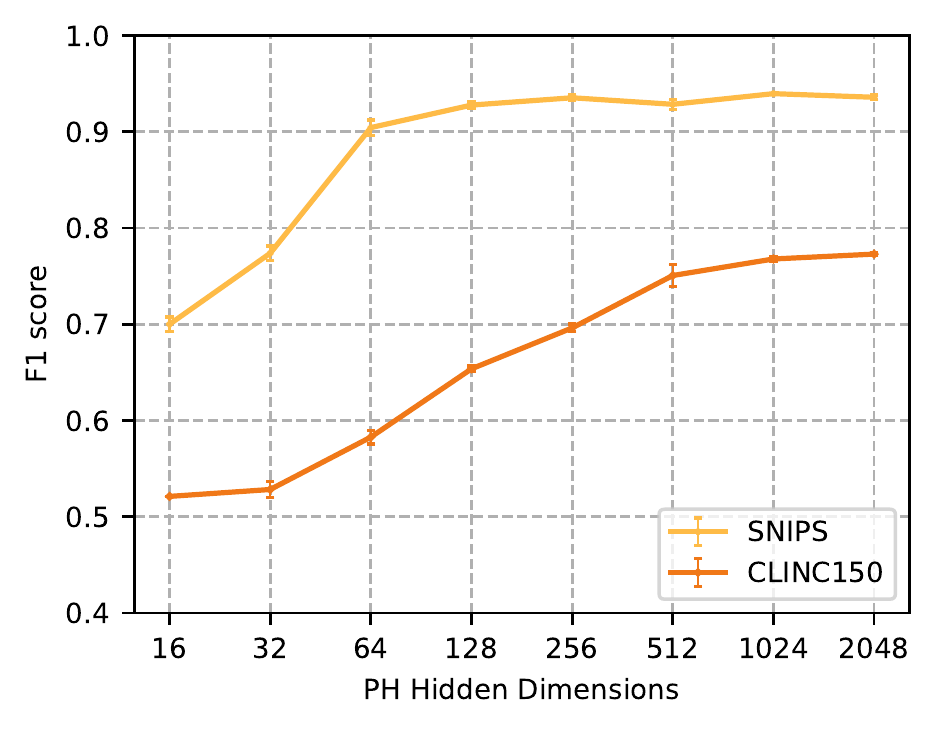}
        \caption{F1 score of PH w.r.t. hidden dimension size}
        \label{fig:f1-vs-hidden-dims}
\end{figure}

Furthermore, larger PHs perform better than smaller PHs but with diminishing returns at higher ends, indicating a sweet spot of PH design.
We observe 512 to be the sweet spot of PH design for SNIPS as it performs better or similar to the other configurations across all experiments.
This finding is corroborated with results on the Clinc-150 dataset.
Figure~\ref{fig:f1-vs-hidden-dims} shows the F1 score of PHs with varying hidden dimensions on both SNIPS and Clinc-150.
We observe similar trends for diminishing return in performance for Clinc as the PH design gets larger.
Similarly, 512 is the inflection point of F1 score improvement, making it the sweet spot PH design for Clinc.
Furthermore, we observe a slower rate of F1 score improvement with respect to hidden dimension size for Clinc than SNIPS.
This can be explained with the observation that Clinc is a more diverse and challenging task with significantly more classes than SNIPS, as described in Section~\ref{subsec:datasets},

\begin{figure}[h]
    \centering
        \includegraphics[width=0.85\columnwidth]{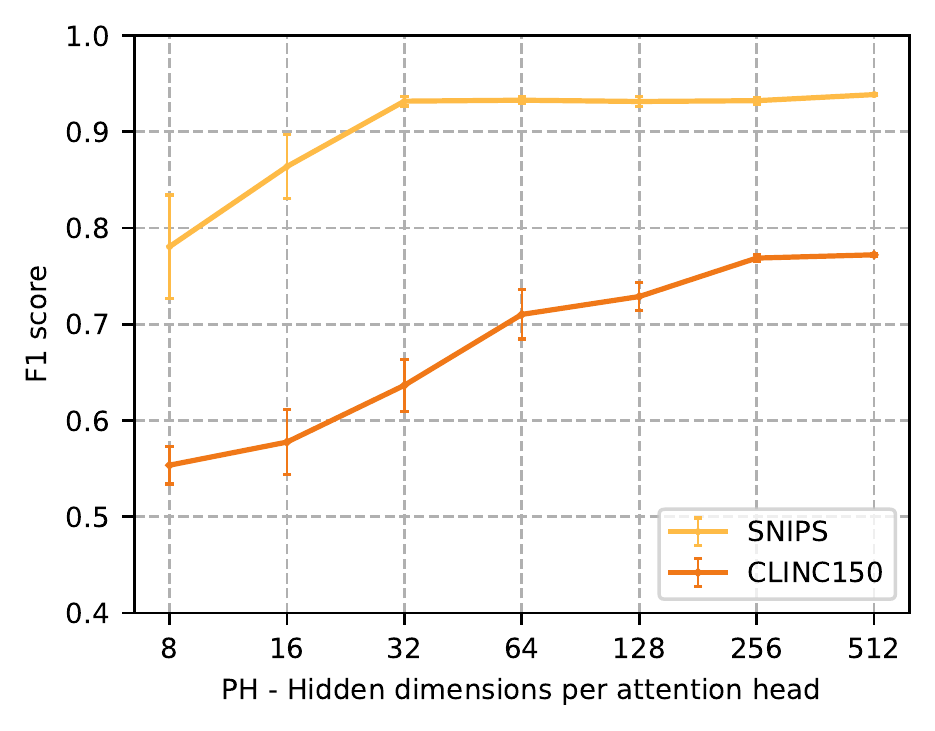}
        \caption{F1 score of PH w.r.t. hidden dims per head}
        \label{fig:f1-vs-dim-per-head}
\end{figure}

\textbf{\# of Attention Heads} We study the impact of attention heads on PHs performance.
Figure~\ref{fig:f1-vs-dim-per-head} shows F1 score w.r.t hidden dimensions per attention head.
We follow the design in~\cite{vaswani2017attention} where the hidden dimensions of the feed-forward layer are effectively distributed evenly among the available attention heads.
We observe that PHs achieve better performance with higher hidden dimensions per head but eventually see diminishing returns.

\subsection{Scalability}
\label{subsec:scalability}
\begin{figure}[h]
    \centering
    \begin{minipage}{.59\columnwidth}
    \centering
        \includegraphics[width=\columnwidth]{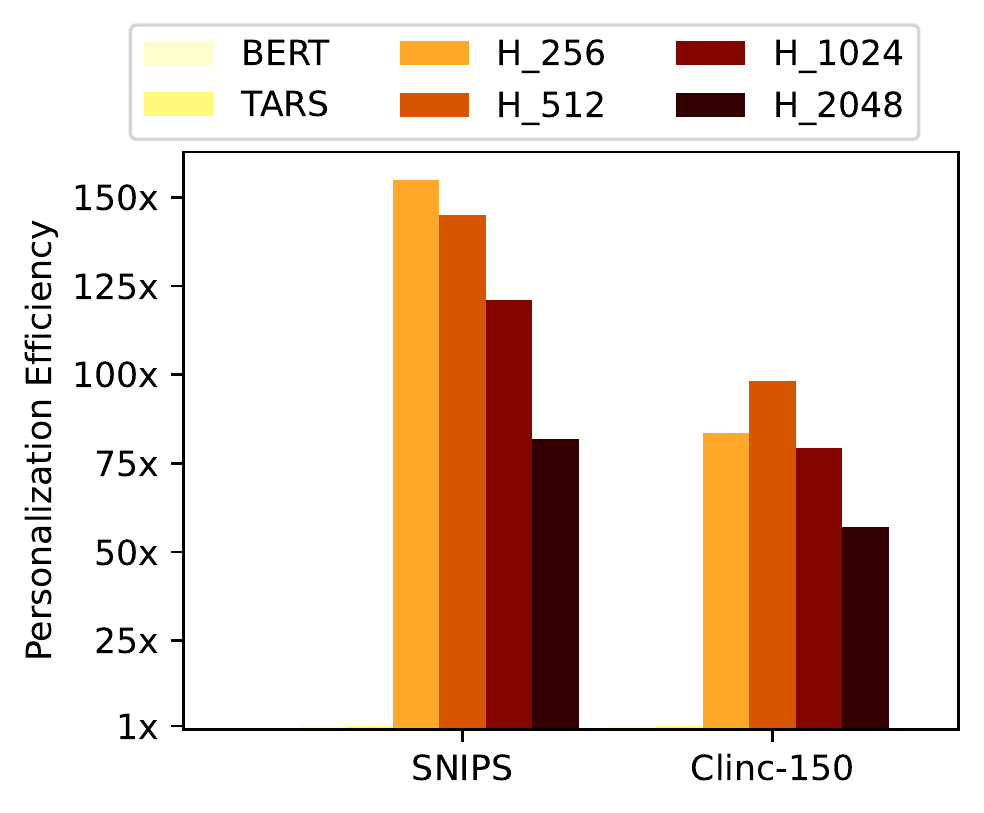}
        \caption{Personalization Efficiency}
        \label{fig:scalability}
    \end{minipage}
    \begin{minipage}{.39\columnwidth}
    \centering
        \includegraphics[width=\columnwidth]{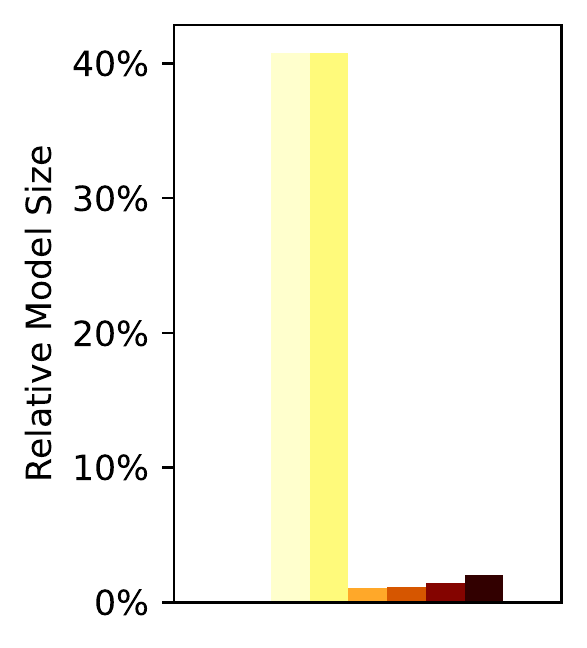}
        \caption{Data Scalability}
        \label{fig:data-scalability}
    \end{minipage}
\end{figure}

We study the scalability of the PH approach and its impact on production deployment.
To help us holistically evaluate a personalization approach, we first introduce a new metric, Personalization Efficiency ($PE$):
\begin{equation}
    PE=\frac{F-score^2}{Training\ Cost\times Model\ Size}
\end{equation}
This new metric considers both the model performance and the computation requirements of training and inference.
We use the number of trainable parameters as an approximation for training cost in this study.
Figure~\ref{fig:scalability} shows the efficiency of 4 PH configurations normalized to the fine-tuning BERT baseline.
We show that PHs achieve efficiency up to 155X compared to the fine-tuning baseline.

Furthermore, for SNIPS we observe smaller PHs generally measure higher in efficiency than larger PHs but see a diminishing return.
For Clinc, 512 achieves the highest efficiency of the PHs tested.
This corroborates our recommendation earlier that 512 is the sweet spot for PH design.

We also quantify the potential storage overhead required for personalized models.
Figure~\ref{fig:data-scalability} shows the additional storage overhead required by the personalized models per individual user relative to the existing user data in production.
We use Gmail as an example application.
We calculate approximately the current per-user data usage based on a report that Gmail user creates up to 1.4MB of data per day and 3 years is the average account lifetime~\cite{backupify-gmail}.
Figure~\ref{fig:data-scalability} shows that the proposed PHs constitute 1\% - 1.5\% of additional storage overhead across all 4 sizes, while the fine-tuning baselines would incur around 40\% additional storage overhead per user.

\section{Related Work}
One of the most common solutions to many natural language understanding problems today is leveraging large-scale pre-trained transformer-based language models~\cite{devlin2018bert, roberta} which are typically trained on language understanding objectives such as Masked Language Modeling and Next Sentence Prediction. These language models are then fine-tuned for a specific task. This transfer of learning to the task of interest is achieved by tuning all model weights on that new task. The performance of these LMs have shown to scale with model size~\cite{scaling}, resulting in massive models consisting of billions of parameters~\cite{gpt3, raffel2020exploring, sanh2021multitask}. When applied to the online setting of personalization training, the applicability of these language models is severely constrained as it results in a dedicated model for each user. This section explores existing works improving the applicability of transformer models at scale.

\subsection{Zero-shot Learning}
Zero-shot learning approaches aim to provide generalized models for a range of language-based tasks without needing additional training steps required by traditional transfer learning approaches. Recent approaches to this problem frame this as a text-to-text generation task~\cite{gpt3, raffel2020exploring, sanh2021multitask} with focuses on prompt design~\cite{perez2021true, khashabi2020unifiedqa}. While shown to be effective in tasks such as QA and summarization, zero-shot performance is still lacking when it comes to text classification~\cite{halder2020task, yinroth2019zeroshot}. This is further shown in our zero-shot experimental results.~\citet{halder2020task} explores the shortcomings of the existing transfer learning mechanisms for text classification, proposing the formalization of text classification as a general binary classification problem. We drew inspiration from this approach for our PH model design.

\subsection{Adapter Networks}
Another method for fine-tune transformers is adapter networks~\cite{houlsby2019parameterefficient, pfeiffer2021adapterfusion}. Adapters are new modules that add a fully connected residual block for each unique downstream task and fine-tune the layer normalization parameters. Our approach is similar in nature but instead applies a small trainable module allowing us to keep the base language model completely frozen during training.
\section{Conclusion}
Today's AI experience remains largely homogeneous across users.
This is because they are often served with the same pre-trained models.
Many applications prefer or even require AI capabilities personalized at the individual level.
In this work, we investigate achieving personalized intelligence at scale.
We introduce a novel model training and inference framework, where a small personalization head is created to adapt large-scale pre-trained LMs.
We only train the PHs and keep the base LM frozen, thus significantly reducing the computation and storage cost compared to the current full model fine-tuning approaches.
We show that the PHs outperform zeroshot models in accuracy and scales significantly better than existing fine-tuning approaches.
We conduct a series of studies to understand the learning behaviors of PHs and provide recommendations on how to design PHs for data and compute efficiency.

\bibliography{anthology,custom}
\bibliographystyle{acl_natbib}
\clearpage
\appendix
\section{Appendix}
\subsection{More PH configurations}

\begin{table}[h]
\centering\scriptsize
\begin{tabular}{c|c|c|c}
                                    \hline 
\textbf{hidden dim} & \textbf{\# attn. heads}  & \textbf{SNIPS}  & \textbf{Clinc} \\
\cline{1-4}\hline
                                 \multirow{3}{*}{64} & 8 & 95.89 & 57.85 \\
                                                     & 4 & 96.32 & 54.41 \\
                                                     & 2 & 96.03 & 56.60 \\
                                \hline
                                 \multirow{3}{*}{32} & 8 & 96.32 & 51.37 \\
                                                     & 4 & 96.24 & 50.83 \\
                                                     & 2 & 95.15 & 46.92 \\
                                \hline
                                 \multirow{3}{*}{16} & 8 & 94.87 & 49.92 \\
                                                     & 4 & 94.43 & 49.99 \\
                                                     & 2 & 94.94 & 50.28 \\
                                 \hline 
                                  \multirow{3}{*}{8} & 8 & 83.09 & 52.58 \\
                                                     & 4 & 80.86 & 52.66 \\
                                                     & 2 & 86.34 & 53.29 \\
                                \hline
\end{tabular}
\caption{F1 score of PHs with smaller sizes.}
\label{tab:small-heads-f1}
\end{table}

Table~\ref{tab:small-heads-f1} shows the F1 score of PHs with 64, 32, 16, and 8 hidden dimensions, on SNIPS and Clinc datasets.
This is an extension to the result shown in Table~\ref{tab:all-f1}.
We observe similar trends carry over to this set of even smaller PHs.
This shows that even a tiny PH can adapt LM well to the SNIPS task.
On the other hand, the smaller PHs are not as effective for the more challenging Clinc datasets.

\subsection{Training for more epochs on Clinc}
Table~\ref{tab:f1-clinc-100} shows the F1 scores of PHs of 4 different sizes on Clinc when training for an additional 50 epochs (100 epochs in total).
This shows that PHs performance continues to improve with more training iterations, indicating that continuing training for more iterations are beneficial in improving PH performance.

\begin{table}[h]
\centering\scriptsize
\begin{tabular}{c|c|c}
                                    \hline 
\textbf{hidden dim} & \textbf{\# attn. heads}   & \textbf{Clinc} \multirow{3}{*}{} \\
\cline{1-2}                                \hline
                                 \multirow{3}{*}{2048}   & 2  & 78.29\\
                                                       & 4  & 77.25 \\
                                                       & 8  & 77.78 \\
								\hline
                                 \multirow{3}{*}{512}   & 2  & 75.82 \\
                                                       & 4  & 75.05 \\
                                                       & 8  & 74.98 \\													   
                                \hline
								\multirow{3}{*}{128} & 2    & 65.29 \\
                                                       & 4  & 65.72 \\
                                                       & 8  & 63.63 \\
								\hline									
                                \multirow{3}{*}{32}   & 2  & 51.35 \\
                                                       & 4  & 53.24 \\
                                                       & 8  & 52.84 \\	
                                \hline 

\end{tabular}
\caption{F1 score of PHs trained on Clinc-150 dataset for an additional 50 epochs (100 epochs in total)}
\label{tab:f1-clinc-100}
\end{table}

\subsection{Analyzing \# of attention heads}
\begin{figure}[h]
        \includegraphics[width=0.8\columnwidth]{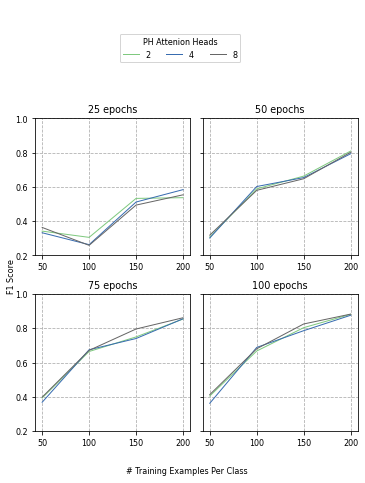}
        \caption{With the same number of training epochs, the impact of training data on performance.}
        \label{fig:attn-heads-constant-epoch}
\end{figure}

\begin{figure}[h!]
        \includegraphics[width=0.8\columnwidth]{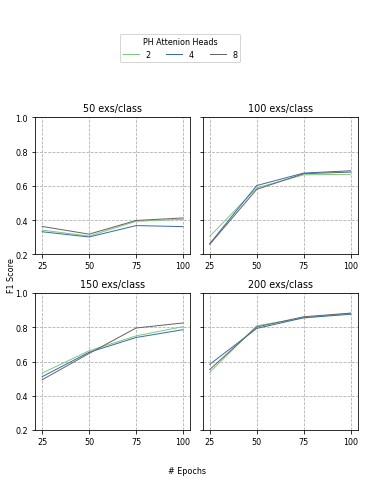}
        \caption{With the same amount of training examples per class, the impact of epochs on performance.}
        \label{fig:attn-heads-constant-data}
\end{figure}

Figure~\ref{fig:attn-heads-constant-epoch} and~\ref{fig:attn-heads-constant-data} analyze the impact of \# attention heads on the performance of PHs.
We conduct experiments similar to that in Section~\ref{subsec:data-vs-epoch}.
We gradually increase the amount of training data while holding the training epochs fixed and measure the F1 score at each stopping point, and vice versa.
We observe that, compared to hidden dimension sizes, \# of attention heads has less effect on the learning behavior and capacity of the PHs.

\end{document}